\documentclass[11pt]{article}
\pdfoutput=1

\usepackage{graphicx}
\usepackage{cite}

\usepackage[in]{fullpage}
\usepackage{float}
\usepackage{amsmath}
\usepackage{array}
\usepackage{color}
\usepackage{enumitem}
\usepackage{hyperref}

\usepackage{float}
\usepackage{amsmath,amssymb,bbm}

\usepackage{algorithm}
\usepackage{algorithmic}

\usepackage{array}
\usepackage{caption}
\usepackage[caption=false,font=normalsize,labelfont=sf,textfont=sf]{subfig}

\newcommand\R{\mathbb{R}}

\usepackage{mathtools}

\newcommand{\beq}{\begin{equation}}
\newcommand{\eeq}{\end{equation}}

\renewcommand{\c}{\mathbf{c}}
\newcommand{\p}{\mathbf{p}}

\renewcommand{\u}{\mathbf{u}}
\renewcommand{\v}{\mathbf{v}}
\newcommand{\y}{\mathbf{y}}
\newcommand{\x}{\mathbf{x}}

\usepackage{url}
\usepackage{amsmath}

\usepackage{authblk}

\author[1]{Konstantina Christakopoulou\thanks{christa@cs.umn.edu}}
\author[2]{Jaya Kawale\thanks{kawale@cs.umn.edu}}
\author[1]{Arindam Banerjee\thanks{banerjee@cs.umn.edu}}
\affil[1]{Department of Computer Science \& Engineering, University of Minnesota, USA}
\affil[2]{Netflix, USA}

\date{}
\begin{document}

\title{Recommendation under Capacity Constraints}

\maketitle
\begin{abstract}

In this paper, we investigate the common scenario where every candidate item for recommendation is characterized by a maximum capacity, i.e., number of seats in a Point-of-Interest (POI) or size of an item's inventory. Despite the prevalence of the task of recommending items under capacity constraints in a variety of settings, to the best of our knowledge, none of the known recommender methods is designed to respect capacity constraints. To close this gap, we extend three state-of-the art latent factor recommendation approaches: probabilistic matrix factorization (PMF), geographical matrix factorization (GeoMF), and bayesian personalized ranking (BPR), to optimize for both recommendation accuracy and expected item usage that respects the capacity constraints. We introduce the useful concepts of user propensity to listen and item capacity. Our experimental results in real-world datasets, both for the domain of item recommendation and POI recommendation, highlight the benefit of our method for the setting of recommendation under capacity constraints. 


\end{abstract}

\section{Introduction}
\label{sec:intro}
Consider what would happen if a Point-of-Interest (POI) recommendation system suggests to a large number of users to visit the same POI, e.g. the same attraction in a theme park, or the same coffee shop; or what would be the effect of an item recommendation system recommending the same products (e.g. movies, jackets) to the vast majority of customers. It is easy to imagine that in the first case the recommended POI might  get overcrowded, resulting in long queues, thus high waiting times. In the second case, the customers might view an `out of stock' or `server overload' message. 

The above scenarios, although seemingly different, share the following key property: every item candidate for recommendation is associated with a maximum capacity -- for a POI it could be the number of visitors allowed at the same time or the number of seats or tables; for a product it could be the maximum number of copies that can be purchased/consumed simultaneously. The task of recommending items under capacity constraints is present in a variety of real-world settings; 
from suggesting the best road for driving while keeping the roads from getting congested, to deciding how to serve personalized content in rush watching periods such as Prime Time or the Oscars. 

However, to the best of our knowledge, none of the available recommendation approaches for item and POI recommendation is designed to respect the capacity constraints. They are often designed to optimize rating prediction \cite{salakhutdinov2011probabilistic, shan2010generalized} or personalized ranking accuracy \cite{christakopoulou2015collaborative, rendle2009bpr, shi2012climf}. For example, latent factor collaborative filtering methods \cite{koren2009matrix}, a popular approach for recommendation, usually optimize only for recommendation accuracy by considering the square loss or other metrics, using the observed user-item ratings. Similar is the case for state-of-the art POI recommendation approaches \cite{li2015rank, lipoint, yu2015survey}. 

In this paper, we present a novel approach that both optimizes recommendation accuracy, measured suitably by rating prediction or personalized ranking losses, and penalizes excessive usage of items that surpasses the corresponding capacities. 
We show how to apply our approach to three state-of-the-art latent factor recommender models: probabilistic matrix factorization (PMF) \cite{salakhutdinov2011probabilistic},  geographical matrix factorization (GeoMF) \cite{lian2014geomf}, and bayesian personalized ranking (BPR) \cite{rendle2009bpr}. We introduce the concepts of \emph{user propensity to listen} (i.e., the probability to listen to the recommendations) and \emph{item capacity}. Both concepts are key factors for estimating the extent to which the expected usage of the items violates the capacity constraints. We present experiments in real-world recommendation datasets that highlight the promise of our approach. 

Our experimental results show that: (i) our formulation allows one to trade-off between optimizing for recommendation accuracy and respecting capacity constraints, 
(ii) alternating minimization empirically works well in practice for learning the model, and (iii) the formulation is reasonably robust to different choices of propensities, capacities, and surrogate loss for the capacity objective. 

The rest of the paper is organized as follows. We briefly review related work in Section \ref{sec:related}. In Section \ref{sec:model} we introduce the concepts of user propensities and item capacities, and devise our method for recommendation under capacity constraints. We empirically evaluate our method in Section \ref{sec:experiments}, and conclude in Section \ref{sec:concl}.

\section{Related Work}
\label{sec:related}
\noindent Matrix factorization (MF) based approaches have become very popular in recommendation systems \cite{koren2009matrix} both for implicit \cite{hu2008collaborative, johnson2014logistic} and explicit feedback \cite{salakhutdinov2011probabilistic}. They predict a user's rating on an item on the basis of low dimensional latent factors, by optimizing square loss \cite{salakhutdinov2011probabilistic, shan2010generalized} or ranking inspired losses \cite{christakopoulou2015collaborative, rendle2009bpr, shi2012climf}. MF approaches have also been employed in the rapidly grown area of POI recommendation where geo-location is used to further improve recommendation performance \cite{lian2014geomf, gao2013exploring, li2015rank, lipoint} -- we refer the reader to \cite{yu2015survey} for a survey of the topic. 

Our work can be viewed in the context of multiple objective optimization models. The latter have been applied in recommendation, either to optimize various ranking metrics \cite{svore2011learning}, to promote diversity \cite{jambor2010optimizing}, or to increase time spent along with click-through-rate \cite{agarwal2012personalized, agarwal2011click}. Our work has also connections with resource-constrained models \cite{Herbrich16}. 
An alternative view of the problem is to minimize the waiting time. Thus queueing theory \cite{karatzas2012brownian} becomes relevant.  

We now briefly review the methods of PMF \cite{salakhutdinov2011probabilistic}, BPR\cite{rendle2009bpr}, and GeoMF \cite{lian2014geomf}, after introducing some useful notation. 

\noindent \textbf{Notation.} Let $N$ be the total number of items, and $M$ the total number of users. Every user is denoted by an index $i=1, \ldots, M$ and every item by $j = 1, \ldots, N$. Assume that user $i$ has rated a subset of the $N$ items, denoted by $\mathcal{L}_i$, and every item has been rated by the set of users $\text{Ra} (j)$. 

\noindent \textbf{PMF.} 
Let $U \in \R^{k \times M}$ be the latent factor corresponding to the users, where the $i^{th}$ column $\u_i \in \R^k$ is the latent factor for user $i$. Similarly, let $V \in \R^{k \times N}$ be the latent factor for the items, where the $j^{th}$ column $\v_j \in \R^k$ is the latent factor for item $j$. Then, \texttt{PMF} models the predicted rating of user $i$ on item $j$ as $\hat{r}_{ij} = \u_i^T \v_j$, so that the overall score matrix $R \in \R^{M \times N}$ is of rank $k \ll M, N$, and thus can be approximated by $U^T V$. 
The objective of \texttt{PMF} \cite{salakhutdinov2011probabilistic} is
\begin{equation*}
\mathcal{E}_{\texttt{PMF}} (U, V) = \sum_{i=1}^{M} \sum_{j \in {L_i}} (r_{ij} - \u_i^T \v_j)^2 + \lambda (\|U\|_F^2 + \|V\|_F^2 ), 
\end{equation*}
where the second term is a L-2 regularization term to prevent over-fitting in the training data, with $\lambda$ denoting the regularization parameter and $\|\cdot\|_F$ the Frobenius norm. 

\noindent \textbf{BPR.} 
BPR \cite{rendle2009bpr} focuses on correctly ranking item pairs instead of scoring single items. Let $L_{i, +}, L_{i, -}$ denote the set of positively (+1) and negatively rated (-1) items by user $i$ respectively. Maximizing the posterior distribution that a user will prefer the positive items over the negative ones, the MF-based BPR objective is: 
\begin{align*}
\mathcal{E}_{\texttt{BPR}} (U, V) &= \sum_{i=1}^{M} \sum_{k \in L_{i,+}} \sum_{j \in {L_{i,-}}} \log (1 + \exp(-\u_i^T (\v_k - \v_j))) \\
& + \lambda (\|U\|_F^2 + \|V\|_F^2 . 
\end{align*} 
%

\noindent \textbf{GeoMF.} 
In POI recommendation, the rating matrix $R$ contains the check-in data of $M$ users on $N$ POIs. Because of the natural characterization of POIs in terms of their geographical location (latitude and longitude) and the spatial clustering in human mobility patterns, recommendations can be further improved \cite{yu2015survey}. This insight has led to the development of \texttt{GeoMF} \cite{lian2014geomf} which jointly models the geographical and MF component. The key idea behind \texttt{GeoMF}, and the related works of  \cite{li2015rank, yu2015survey}, is that if a user $i$ has visited a POI $j$  but has not visited the nearby POIs, these nearby POIs are more likely to be disliked by this user compared to far-away non-visited POIs. \texttt{GeoMF} represents the users and items not only by their latent factors $U, V$, but also by the activity and influence vectors $X, Y$ respectively. By splitting the entire world into $L$ even grids, \texttt{GeoMF} models a user $i$'s \emph{activity area} as a vector $\x_i \in \R^{L}$, where every entry $x_{i\ell}$ of the vector denotes the possibility that this user will appear in the $\ell$-th grid. Similarly, the model represents every POI by a vector $\y_j \in \R^{L}$, referred to as \emph{influence vector}, where every entry $y_{j\ell}$ indicates the quantity of influence POI $j$ has on the $\ell$-th grid. 
While the activity area vector $\x_i$ of every user $i$ is latent, the influence vectors $\y_{j}$ for every POI $j$ are given as input to the model and are pre-computed using kernel density estimation \cite{zhang2013igslr} as follows. The degree of the influence POI $j$ has on the $\ell$-th grid is computed as 
\label{eq:kernel}
$y_{j \ell} = \frac{1}{\sigma} \mathcal{K} (\frac{d(j, \ell)}{\sigma})$
where $\mathcal{K}$ is the standard Gaussian distribution, $\sigma$ is the standard deviation and $d(j, \ell)$ is the Euclidean distance between the POI $j$ and the $\ell$-th grid. Although in \cite{lian2014geomf} the activity area vectors are constrained to be non-negative and sparse, in our work we do not make such an assumption. Concretely, \texttt{GeoMF} predicts user $i$'s rating on POI $j$ as $\hat{r}_{ij} = \u_i^T \v_j + \x_i^T \y_j$. \texttt{GeoMF} uses the point-wise square loss of \texttt{PMF}; however, we can also replace this with the \texttt{BPR} loss, giving rise to a method we refer to as \texttt{Geo-BPR}.

\section{Proposed Framework}
\label{sec:model}
While \texttt{PMF},  \texttt{GeoMF} and \texttt{BPR} focus on accurately predicting the best items to be recommended to every user, they do not consider any capacity constraints that might be associated with the items. 
However, for POI (e.g. theme parks attractions or coffee shops) recommendation, if the system recommends POIs overlapping across the majority of users, the users who follow the recommendation might have to wait in long queues, although there might be perfectly good POIs empty just around the corner. 
This scenario and more (e.g. viral video recommendation, limited shelf on a virtual store, road congestion) can benefit from a system that gives recommendations while respecting the capacity constraints. 

In the following, we devise our method referred to as \texttt{Cap-PMF}/ \texttt{Cap-BPR} for item recommendation, and \texttt{Cap-GeoMF}/\texttt{Cap-GeoBPR} for POI recommendation. The key idea of our approach is that we want to both optimize for recommendation accuracy (accurately predicting user ratings / accurately ranking items) and penalize when the items'  expected usage exceeds the respective capacities. \\


\noindent \textbf{User Propensities.} 
In our approach, every user $i$ is associated with a variable, indicating the probability that he will listen to the system's recommendations. We refer to this variable as \emph{propensity to listen} and denote it with $p_i \in [0,1]$, resulting in a vector of propensities $\p \in \R^{M}$ for all $M$ users. 
There are many ways propensities can be modeled; one possible definition is: 
\begin{equation*}
\label{eq:defprop}
p_i = \frac{\text{\# times user $i$ followed the recommendation}}{\text{\# user $i$-system interactions}}, 
\end{equation*}
where feedback of the form user follows/ignores the recommendation is needed. We explore different ways of setting the user propensities in Section \ref{subsec:setting}. 

Propensity to listen to the recommendations can be an inherent user property: some users tend to listen to the system's recommendations more, compared to others. 
However, the same user's propensity might vary with time, e.g. 
the user might go to a certain place for lunch every day at 2pm, so no matter how good the restaurant recommendation is, he will not listen (low propensity); in contrast, he might be experimental during his dinner time (high propensity). Also, factors such as who the user is with or quality of experience with the system can affect the user's propensity. 
For such cases we plan to include dynamic propensity estimation in the future. 
We argue that user propensity, which has connections to the themes of \cite{liang2016modeling, schnabel2016recommendations}, is a key factor for recommendation systems. Although here we use it to compute the expected usage of items, one can also use it to e.g. target the users with low/high propensities.\\ 
\noindent \textbf{Item Capacities.} Every item $j=1,\ldots,N$ is characterized by a parameter indicating the maximum number of users who can simultaneously use it. We refer to this variable as \emph{capacity} and denote it with $c_j>0$, resulting in a vector of capacities $\c \in \R_{+}^{N}$ for all $N$ items. 
For POI recommendation, a POI's capacity could be the total number of seats, or number of visitors allowed per time slot. For general item recommendation, an item's capacity could be the maximum number of users that can watch the same movie without leading to a system crash, or the maximum number of copies of the same item in the inventory.  We explore different ways of setting the item capacities in Section \ref{subsec:setting}.



Capacity is key for recommendation systems, as when many users are directed to the same item, the item will quickly reach its capacity. This will in turn lead to deteriorated user experience, such as long waiting times or out of stock items. This motivates the need for a recommendation system that respects the items' capacities. \\

\noindent \textbf{Expected Usage.} We define the expected usage of an item $j$ as the expected number of users who have been recommended item $j$ and will follow the recommendation:
$
\sum_{i=1}^{M} p_i \hat{r}_{ij}, 
$
where $\hat{r}_{ij}$ denotes the predicted rating of user $i$ on item $j$. 
If $\hat{r}_{ij}$ was either 1 or 0, the $\sum_{i=1}^{M} \hat{r}_{ij}$ term would indicate the total number of users who have been recommended item $j$. Now, $\hat{r}_{ij}$ is given by
\begin{equation}
\label{eq:rhat}
\hat{r}_{ij}=
 \left\{\begin{aligned}
        &\u_i^T \v_j ~~~~~~~~~~~~~ \text{for \texttt{Cap-PMF}}\\
        &\u_i^T \v_j + \x_i^T \y_j ~~ \text{for \texttt{Cap-GeoMF}}~.
       \end{aligned}
 \right.
\end{equation}
For ease of optimization, we do not threshold $\hat{r}_{ij}$ to be either 0 or 1. Instead, we constrain $\hat{r}_{ij}$ in the range of [0, 1] by using the sigmoid function, $\sigma (\cdot) = \frac{1}{1+ \exp(-(\cdot))}$, leading to: \beq
\label{eq:exp-usage2}
\mathbf{E} [\text{usage}(j)] = \sum_{i=1}^{M} p_i \sigma(\hat{r}_{ij})~, 
\eeq 
which is the weighted combination of the estimated ratings for the users, using as weights the corresponding user propensities
. 

Importantly, any  model estimating whether a user $i$ will buy/visit item/POI $j$ can be used to replace Equation \eqref{eq:rhat}, thus leading to a family of recommendation under capacity constraints algorithms.\\\\
\noindent \textbf{Capacity Loss.}
Since we wish to penalize the model for giving recommendations which result in  the expected usage of the items exceeding the corresponding capacities, we want to minimize the average capacity loss, given by:
\beq
\frac{1}{N} \sum_{j=1}^{N} \mathbf{1}[c_j \leq \mathbf{E}[\text{usage}(j)]  ]~.
\label{eq:cap-loss}
\eeq
The use of the indicator function $\mathbf{1}[\cdot](j)$ is not suitable for optimization purposes. We will use a surrogate for the indicator function. Considering the difference $\Delta (c_{j}, \mathbf{E}[\text{usage}(j)])$ $= c_{j} - \mathbf{E}[\text{usage}(j)] $, 
we use the logistic loss of the difference as the surrogate: 
\beq
\ell (\Delta (c_j, \mathbf{E}[\text{usage}(j)])) = \log(1 + \exp(-\Delta (c_j , \mathbf{E}[\text{usage}(j)]))~, 
\label{eq:log-diff}
\eeq
noting that it forms a convex upper bound to the indicator function. Alternative surrogate losses to consider are:
\begin{equation}
\label{eq:losses}
\ell (\Delta (c_{j}, \mathbf{E}[\text{usage}(j)]))=
 \left\{\begin{aligned}
        &\exp(-\Delta ) ~~~ \text{ (Exponential loss)}\\
        & \max(-\Delta,0) ~~ \text{ (Hinge Loss)}
       \end{aligned}
 \right.
\end{equation}
Although the square loss $(-\Delta )^2 $ is a convex surrogate of the indicator, it is not a suitable surrogate for the capacity loss, as it penalizes both positive and negative differences, whereas we want to penalize only the cases when the expected usage \emph{exceeds} the capacity. \\\\
\noindent \textbf{Overall Objective.}
Putting all the pieces together, using logistic as the surrogate loss, our method minimizes the following objective for item recommendation (\texttt{Cap-PMF}): \begin{align}
\label{eq:12}
& \mathcal{E}_{\text{\texttt{cap-PMF}}} (U, V) = (1 - \alpha) \cdot \sum_{i=1}^{M} \sum_{j \in L_i} (r_{ij} - \u_i^T \v_j)^2 \\
\nonumber &+ \alpha \cdot \frac{1}{N} \sum_{j=1}^{N} \log\left(1 + \exp\left( \sum_{i=1}^{M} p_i \sigma(\u_i^T \v_j) - c_{j} \right)\right) + \lambda (\|U\|_F^2 + \|V\|_F^2)~, 
\end{align}
whereas for POI recommendation under constraints our method \texttt{Cap-GeoMF} minimizes:
\begin{align}
\label{eq:13}
& \mathcal{E}_{\text{\texttt{cap-GeoMF}}} (U, V, X) = (1 - \alpha) \cdot \sum_{i=1}^{M} \sum_{j \in L_i} (r_{ij} - \u_i^T \v_j - \x_i^T \y_j)^2,  \\
\nonumber &+ \alpha \cdot \frac{1}{N} \sum_{j=1}^{N} \log\left(1 + \exp\left(   \sum_{i=1}^{M} p_i (\sigma(\u_i^T \v_j + \x_i^T \y_j)) - c_{j} \right)\right)  \\
\nonumber & + \lambda (\|U\|_F^2 + \|V\|_F^2 + \|X\|_F^2)~. 
\end{align}

In the above formulation, $\alpha$ is a fixed, user-chosen parameter in [0, 1] that handles the trade-off between the first part of the objective, i.e., recommendation accuracy as measured by the average rating prediction error, and the second part of the objective, referred to as the \emph{capacity loss}. When $\alpha $ is equal to 0, our model reduces to \texttt{PMF} (or \texttt{GeoMF}). When $\alpha $ is 1, we are not interested in good prediction accuracy -- our only concern is to ensure that every item's expected usage will not exceed its fixed capacity.

If we replace the first objective in \eqref{eq:12} (or \eqref{eq:13}) with the one of BPR to optimize for ranking accuracy, we obtain \texttt{Cap-BPR} (or \texttt{Cap-GeoBPR}): 
\begin{align}
\label{eq:capbpr}
& \mathcal{E}_{\text{\texttt{cap-BPR}}} (U, V) = (1 - \alpha) \cdot \sum_{i=1}^{M} \sum_{k \in L_{i, +}} \sum_{j \in L_{i, -}} \log (1 + \exp(- \u_i^T (\v_k - \v_j))) \\
\nonumber &+ \alpha \cdot \frac{1}{N} \sum_{j=1}^{N} \log\left(1 + \exp\left( \sum_{i=1}^{M} p_i \sigma(\u_i^T \v_j) - c_{j} \right)\right) + \lambda (\|U\|_F^2 + \|V\|_F^2)~, 
\end{align}

Following we devise the updates for \texttt{Cap-PMF} and \texttt{Cap-GeoMF}, but we will experiment in Section \ref{sec:experiments} with all four proposed approaches \texttt{Cap-PMF},  \texttt{Cap-GeoMF}, \texttt{Cap-BPR}, \texttt{Cap-GeoBPR}.

The optimization for \eqref{eq:12}, \eqref{eq:13} is done by alternating minimization, i.e., updating $U$ while keeping $V$ (and $X$) fixed, then updating $V$ while keeping $U$ (and $X$) fixed (and updating $X$ while keeping $U$ and $V$ fixed for \texttt{Cap-GeoMF}). The latent factor updates  are done using gradient descent, and for iteration $t+1$ are:
\begin{align*}
& \forall i=1, \ldots,M, ~~ \u_i^{t+1} \leftarrow \u_i^t - \eta \nabla_{\u_i} \mathcal{E}_{\texttt{cap-MF}} (U^{t}, V^{t}, X^{t}) \\
& \forall j=1,\ldots,N, ~~ \v_j^{t+1} \leftarrow \v_j^t - \eta \nabla_{\v_j} \mathcal{E}_{\texttt{cap-MF}} (U^{t+1}, V^{t}, X^{t})\\
& \forall i=1, \ldots,M, ~~ \x_i^{t+1} \leftarrow \x_i^t - \eta \nabla_{\x_i} \mathcal{E}_{\texttt{cap-GeoMF}} (U^{t+1}, V^{t+1}, X^{t})
\end{align*}
where the last gradient update is valid only for \texttt{cap-GeoMF}. We denote with $\mathcal{E}_{\texttt{cap-MF}}$ either $\mathcal{E}_{\texttt{Cap-GeoMF}}$ or $\mathcal{E}_{\texttt{cap-PMF}}$.
The gradients can be obtained by a direct application of the chain rule.
From Equations \eqref{eq:rhat}, \eqref{eq:exp-usage2} and \eqref{eq:log-diff}, we get the gradients
\begin{align}
& \nabla_{\Delta} \ell(\Delta(c_j, \mathbf{E}[\text{usage}(j)])) = -\sigma(-\Delta(c_j, \mathbf{E}[\text{usage}(j)])) \label{eq:14} \\
& \nabla_{\u_i} \Delta(c_j, \mathbf{E}[\text{usage}(j)]) = -p_i \v_j \sigma(\hat{r}_{ij})\sigma(-\hat{r}_{ij}) \label{eq:15}\\
& \nabla_{\v_j} \Delta(c_j, \mathbf{E}[\text{usage}(j)]) = -\sum_i p_i \u_i \sigma(\hat{r}_{ij})\sigma(-\hat{r}_{ij}) \label{eq:16} \\
& \nabla_{\x_i} \Delta(c_j, \mathbf{E}[\text{usage}(j)]) = -p_i \y_j \sigma(\hat{r}_{ij})\sigma(-\hat{r}_{ij})~, \label{eq:17}
\end{align}
where for \eqref{eq:15}, \eqref{eq:16}, \eqref{eq:17} we used the property $\nabla_x\sigma(x) = \sigma(x)\cdot \sigma(-x)$. 
%
Using Equations \eqref{eq:14}-\eqref{eq:17}, we obtain the gradient of the latent parameters as follows. The gradient of the objective, w.r.t $\u_i$ is
\begin{align*}
& \nabla_{\u_i} \mathcal{E}_{\texttt{cap-MF}} = -(1-\alpha) \sum_{j \in L_i} 2 (r_{ij} - \u_i^T \v_j ) \v_j +  2\lambda \u_i + \\
& +\frac{\alpha}{N} \sum_{j=1}^{N}\sigma(-\Delta(c_j, \mathbf{E}[\text{usage}(j)])) \cdot p_i \v_j \sigma(\hat{r}_{ij})\sigma(-\hat{r}_{ij})~.
\end{align*}
Similarly,  the gradient of the objective, w.r.t $\v_j$ is
\begin{align*}
& \nabla_{\v_j} \mathcal{E}_{\texttt{cap-MF}} = -(1-\alpha) \sum_{i \in \text{Ra}(j)} 2 (r_{ij} - \u_i^T \v_j ) \u_i +  2\lambda \v_j + \\
& +\frac{\alpha}{N} \sigma(-\Delta(c_j, \mathbf{E}[\text{usage}(j)])) \cdot \sum_{i=1}^{M} p_i \u_i \sigma(\hat{r}_{ij})\sigma(-\hat{r}_{ij})~.
\end{align*}
For \texttt{cap-GeoMF}, the gradient of the objective, w.r.t $\x_i$ is
\begin{align*}
& \nabla_{\x_i} \mathcal{E}_{\texttt{cap-GeoMF}} = -(1-\alpha) \sum_{j \in L_i} 2 (r_{ij} - \u_i^T \v_j - \x_i^T \y_j ) \y_j \\
& +  2\lambda \x_i +\frac{\alpha}{N} \sum_{j=1}^{N}\sigma(-\Delta(c_j, \mathbf{E}[\text{usage}(j)])) \cdot p_i \y_j \sigma(\hat{r}_{ij})\sigma(-\hat{r}_{ij})~.
\end{align*}

We have demonstrated here our method in the batch setting. In the future, we will consider the online or bandit setting \cite{mairal2010online, agarwal2009explore}. 

\section{Experimental Results}
\label{sec:experiments}
In this section, we present empirical results with the goal of addressing the following research questions:

\begin{enumerate}
\item What is the interplay of rating prediction, capacity loss and overall objective for \texttt{CapMF} as we vary the capacity trade-off parameter $\alpha$ (Section \ref{subsec:q1})? Similarly for pairwise ranking loss versus capacity loss for \texttt{Cap-BPR} (Section \ref{subsec:ranking})?   
\item How do the proposed approaches \texttt{Cap-PMF}, \texttt{Cap-GeoMF}, \texttt{Cap-BPR}, \texttt{Cap-GeoBPR}  compare with their state-of-the-art unconstrained counterparts? (Section \ref{subsec:q2})
\item Which surrogate loss is the best suited to model the capacity loss? (Section \ref{subsec:q3})
\item How do different capacity vector definitions affect the algorithm's performance? (Section \ref{subsec:capacities})
\item How do different propensity vector definitions affect the algorithm's performance? (Section \ref{subsec:propensities})
\item How do our methods perform for top-N recommendation? (Section \ref{subsec:q6})
\item How do our methods compare with a simple post-processing baseline we introduce for capacity constrained recommendation? (Section \ref{subsec:baseline})
\end{enumerate}

\begin{figure*}[!t]
\subfloat[Foursquare, propensities]{\includegraphics[width=0.25\textwidth]{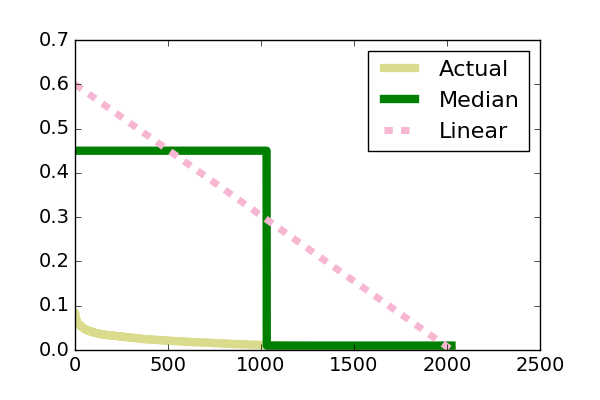}%
}
\subfloat[Gowalla, propensities]{\includegraphics[width=0.25\textwidth]{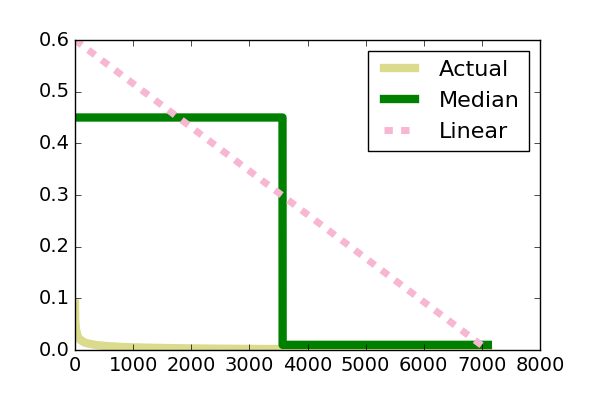}%
}
\subfloat[Foursquare, capacities]{\includegraphics[width=0.25\textwidth]{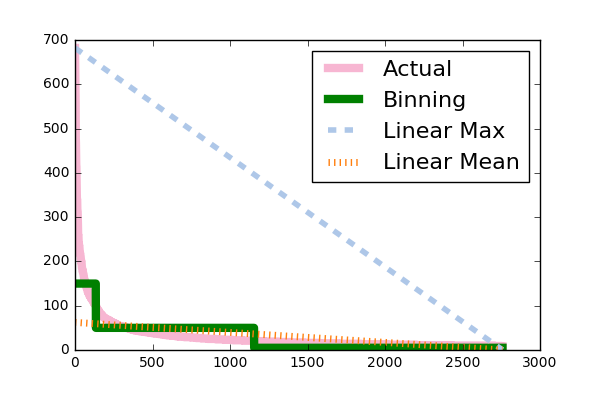}%
}
\subfloat[Gowalla, capacities]{\includegraphics[width=0.25\textwidth]{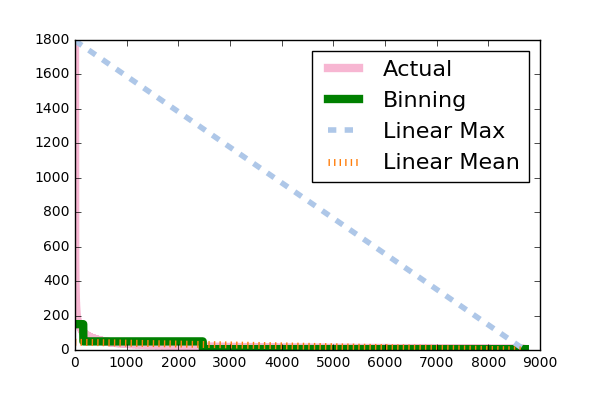}%
}
\caption{Item capacity scores sorted in decreasing order for the various choices of capacities.}
\label{capacities}
\end{figure*}
\subsection{Experimental Setting}
\label{subsec:setting}


\noindent \textbf{Data.} For item recommendation, we considered two public real-world datasets containing users' ratings on movies in a multi-relevance scale of [1, 5]: Movielens 100K and Movielens 1M\footnote{http://www.grouplens.org}. For POI recommendation, we considered two public real-world datasets containing user check-ins in POIs: Gowalla and Foursquare\footnote{http://www.ntu.edu.sg/home/gaocong/data
/poidata.zip}. 
For the Foursquare and Gowalla datasets, we removed the users and POIs with less than or equal to 10 ratings. The users, items and ratings statistics of the data are found in Table \ref{table:1}.

Foursquare and Gowalla contain implicit check-in data (only ratings of 1 and 0 are present). A rating of 1 denotes that the user has visited the POI while a 0 denotes that the user has not visited it; because the user potentially does not like the POI or does not know about it.  
Movielens 100K and Movielens 1M contain ratings in a multi-relevance scale of 1 to 5 stars. We experimented with two feedback setups for these datasets: 
(i) two-scale explicit feedback (+1, -1), using 4 as the threshold of liking, and 
(ii) implicit feedback, marking every non-zero rating as a 1, and keeping the rest as 0.  



\begin{table}[!t]
\caption{Real Data Statistics}
\centering
\begin{tabular}{l|c|c|c}
\hline
Dataset & \# Users & \# Items & \# Ratings \\
\hline
Movielens 100K &  943 & 1,682  & 100,000 \\
Movielens 1M & 6,040 & 3,706  & 1,000,209 \\
Foursquare & 2,025 & 2,759  & 85,988 \\
Gowalla & 7,104  & 8,707  & 195,722 \\
\hline
\end{tabular}
\label{table:1}
\end{table}

\noindent \textbf{Evaluation Setup. } 
We repeated all experiments 5 times by drawing new train/test splits in each round. We randomly selected half of the ratings for each user in the training set and moved the rest of the observations to the test set. This scheme was chosen to simulate the real recommendation setup, where there exist users with a few ratings as well as users with many ratings. 

For the implicit feedback datasets, as only positive observations are available, we introduced negative observations (disliked items) in the training set as follows. For every user with $N_i^{\text{train}}$ positive observations (+1), we sampled $N_i^{\text{train}}$ items as negatives (-1) from the set of items marked as 0 both for the training and the test set. This is common practice to avoid skewed algorithm predictions resulting from training on only positive observations, and to avoid computational overhead of considering all unrated items as negative observations.\footnote{Several strategies for sampling negative user-item pairs are discussed in  \cite{pan2008one}.}. \\



\noindent \textbf{Capacity \& Propensity Setting.} As the information of items' capacities and users' propensities is not available in the considered data, and to the best of our knowledge to any of the publicly available recommendation datasets, in our experiments we considered different ways of setting them. 
For setting capacities, we considered the following three cases: (i) item capacities are analogous to usage, i.e., such a setting is inspired by the supply-demand law: the more users ask for an item, the more copies of the item will be in the market, (ii) item capacities are inversely proportional to usage, i.e., capturing the case when items with low capacities are often in high demand, and (iii) item capacities are irrespective of usage. In particular, we instantiated the above concepts with the following:
\begin{enumerate}[leftmargin=0cm,labelwidth=\itemindent,labelsep=0cm,align=left]
\item `actual': $ \forall j,~ c_j = \text{\# users who have rated item } j$\footnote{An alternative would be \# users who have liked the item.}
\item `binning': Transform actual capacities to the bins: $[0,20] \rightarrow 5$, $[21,100] \rightarrow 50$, $[101, \text{max capacity}]$ $\rightarrow 150$. 
\item `uniform-k': Set all item capacities to a value $k$, e.g. 10. 
\item `linear max': Spread the items capacities in [0, maximum actual capacity] using a linear function. 
\item `linear mean'. Same as `linear max', but spread in the range [0, 2*mean of actual capacities]. 
\item `reverse binning': Map the actual capacities to the following bins (reversely proportional to usage): $[0,20] \rightarrow 150$, $[21,100] \rightarrow 50$, $[101, \text{max capacity}]$ $\rightarrow 5$.
\end{enumerate}
Note that (1) -(2) are analogous to usage, (3) - (5) are irrespective of usage and (6) is proportional to usage. 
Figures \ref{capacities}(c), (d) show  all items' capacity scores sorted decreasingly for the various capacity choices, for the POI datasets. 

For setting user propensities, we considered the following cases: (i) user propensities are analogous to system usage by the user, i.e., capturing the intuition that the more the user interacts with the system the more they tend to listen to the recommendations, and (ii) user propensities are irrespective of usage, i.e., capturing that propensities are an inherent user property. In particular, we considered the following: 
\begin{equation*}
\text{\noindent (1) `actual':~~~~} p_i = \frac{\text{\# observed ratings for user $i$}}{\text{Total \# items}} = \frac{|L_i|}{N}
\end{equation*}
where $|\cdot|$ denotes a set's cardinality.\footnote{An alternative could be: $p_i = \frac{\text{\# liked items for user $i$}}{\text{\# observed ratings for user $i$ }} = \frac{|L_i^{+}|}{|L_i|}$} \\
(2) `median': Set propensities $\geq$ the median of the actual propensities to 0.45, and propensities $<$ median to 0.01. This illustrates two distinct groups of users; those who tend to listen to the system's recommendations, and those who do not. \\
(3) `linear': Spread the user propensities in the range [0, 0.6] using a linear function. \\
Note that (1)-(2) are analogous to user usage, and (3) is irrespective of usage. Figures \ref{capacities}(a), (b) show all users' propensity scores sorted decreasingly for the various propensity choices for the POI datasets. 


%
%
\begin{figure}[ht]
\subfloat[Foursquare, lat./ long. coordinates]{\includegraphics[scale=0.25]{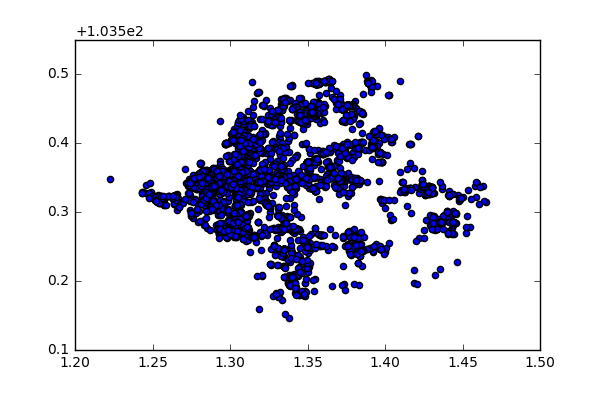}%
}~~
\subfloat[Foursquare, tile coordinates]{\includegraphics[scale=0.25]{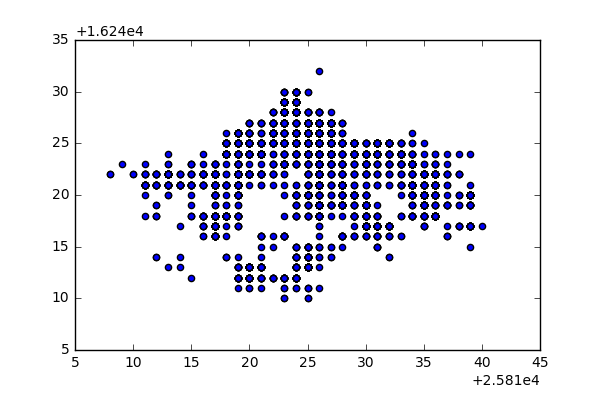}}~~
\subfloat[Gowalla, lat./ long. coordinates]{\includegraphics[scale=0.25]{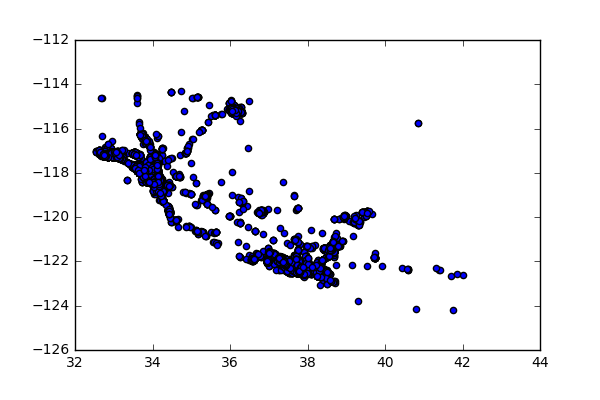}%
}~~
\subfloat[Gowalla, tile coordinates]{\includegraphics[scale=0.25]{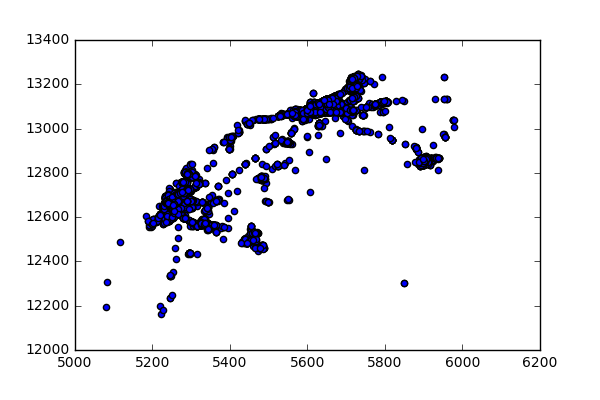}}
\caption{Location information Scatter plots. 
}
\label{locationXY}
\end{figure}
\noindent \textbf{Location.} The POI datasets Foursquare and Gowalla, contain apart from the check-in observations, location information about where every POI lies in terms of geographical latitude and longitude coordinates. In Figures \ref{locationXY}(a), (c) we show scatter plots of the latitudes (x axis) and longitudes (y axis) for the POIs of the non-subsampled datasets of Foursquare and Gowalla respectively.
Using the Mercator projection, and fixing the ground resolution to level of detail set to 15, we transformed the latitude/longitude coordinates of every POI to one of the $2^{15}$ tiles where the entire Earth can be divided into\footnote{https://msdn.microsoft.com/en-us/library/bb259689.aspx}. We show the tiles x-y coordinates of the various POIs for Foursquare in Figure \ref{locationXY}(b) and for Gowalla in Figure \ref{locationXY}(d). We set $L$, i.e., the dimension of the influence vector, to the number of unique tiles found. For Foursquare $L$ is 290, whereas for Gowalla $L$ is 4320. Considering the set of all $L$ unique tiles and representing every POI as a pair of (tileX, tileY) coordinates, we pre-computed the influence matrix $Y \in \mathbf{R}^{L \times N}$ using Kernel Density Estimation \cite{zhang2013igslr}. 
\noindent \textbf{Parameter Setting.} We set the algorithm parameters as follows. Similarly to \cite{lipoint}, we set the rank of the latent factors $k$ to 10. We used for the learning rate of gradient descent the Adagrad rule \cite{duchi2011adaptive}, which for the latent factor of the user $i$, $\u_i$ at iteration $t$ is: $\eta_t = 1/ \sqrt{\sum_{\tau=0}^{t-1} \nabla_{\u_{i, \tau}}^2}$. We fixed the regularization parameter $\lambda$ to 0.00001. 
In practice, to further improve the algorithm's performance, we can tune the parameters of rank and regularization in a validation set. We stopped the training of the algorithm either when the improvement in the value of the optimization objective in the training set was smaller than  0.00001 or after 3,000 iterations. 

\noindent \textbf{Evaluation Metrics.} We report the method's test-set performance in terms of the following metrics: 

(1) For \texttt{CapMF}, we report \emph{Root Mean Square Error (RMSE)} which measures test set rating prediction accuracy:
\beq
\text{RMSE } = \sqrt{\frac{1}{M} \sum_{i=1}^{M} \frac{1}{|L_{i, \text{test}}|}\sum_{j \in L_{i, \text{test}}}(\hat{r}_{ij} - r_{ij})^2}~.
\eeq $L_{i, \text{test}}$ denotes the set of observed ratings for user $i$ in the test set. \\
(2) For \texttt{Cap-BPR} we report \emph{0/1 Pairwise Loss} which measures the average number of incorrectly ordered pairs (-1 ranked above +1):
\beq
\text{0/1 Pair. Loss} = \frac{1}{M}\sum_{i=1}^{M} \frac{1}{|L_{i}^{-}|\cdot |L_{i}^{+}|}\sum_{j \in L_{i, \text{test}}^{-}} \sum_{k \in L_{i, \text{test}}^{+}} \mathbf{1}[\hat{r}_{ij} \geq \hat{r}_{ik}], 
\eeq where $L_{i, \text{test}}^{+}$ and $L_{i, \text{test}}^{-}$ denote the set of positively and negatively rated items by user $i$ in the test set respectively. \\
(3) \emph{Capacity Loss}, which measures on average the extent to which the recommendations lead to violating the capacity constraints:
\beq
\text{Capacity Loss } = \frac{1}{N} \sum_{j=1}^{N} \ell\left(c_{j} - \sum_{i=1}^{M} p_i \sigma(\hat{r}_{ij}) \right)~.
\eeq  
For metrics (1), (2), (3) values closer to 0 are better. \\
(4) \emph{Overall Objective}. Given that our methods optimize the two objectives of rating prediction (\texttt{CapMF})/ranking (\texttt{Cap-BPR}) loss and capacity loss, we also report:
$(1-\alpha) \text{RMSE}^2 + \alpha \text{Capacity Loss} $ for \texttt{CapMF} (or $(1-\alpha) \text{0/1 Pairwise Loss} + \alpha \text{Capacity Loss} $ for \texttt{Cap-BPR}). \\
(5) \emph{Mean Average Precision @ k}, to measure top-k recommendation quality. AP@k is the average of Precisions computed at each relevant position (+1) in the top $k$ items of the user's ranked list. Precision@k (P@k) equals the fraction of relevant items out of the top k items. 
\beq
AP@k = \sum_{r=1}^{k} \frac{P@r \cdot rel(r)}{\min(k, \text{\# relevant items})},  
\eeq where $rel(r)$ is 1 if the item in position $r$ is relevant, and 0 otherwise. 
After computing Average Precision per user, we average the results over all users. Values closer to 1 are better. 

\noindent \textbf{Baselines. } 
We compare our proposed methods with the baselines of: (i) \texttt{PMF} \cite{salakhutdinov2011probabilistic}, (ii) \texttt{BPR} \cite{rendle2009bpr} (item recommendation), (iii) \texttt{GeoMF} \cite{lian2014geomf}, (iv) \texttt{Geo-BPR} (POI recommendation), and (v) \texttt{onlyCap}, i.e., the baseline of setting $\alpha$ to 1.  Although no method has been proposed for recommendations under capacity constraints in the literature, in \ref{subsec:baseline} we introduce a simple post-processing baseline.

\begin{figure*}[ht]
\centering
\subfloat[Movielens 100K]{\includegraphics[width=0.25\textwidth]{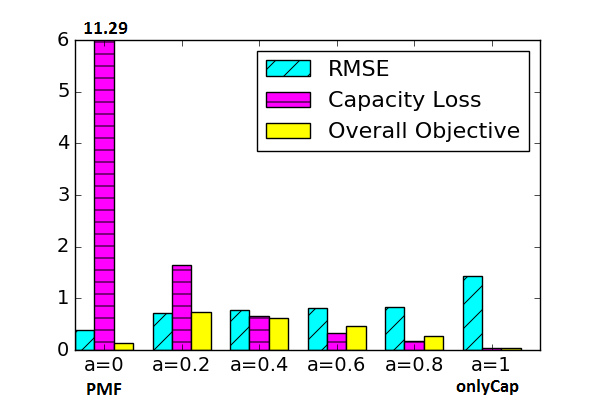}%
}
\subfloat[Movielens 1M]{\includegraphics[width=0.25\textwidth]{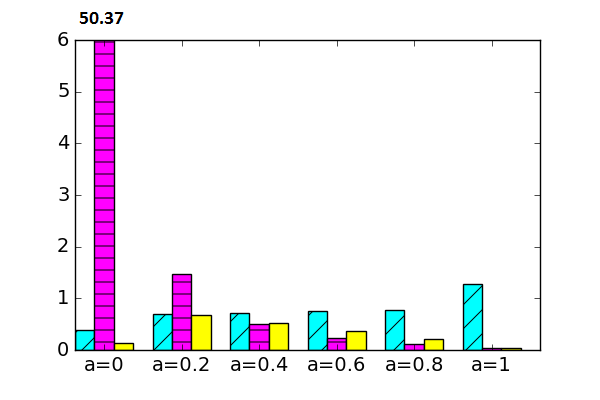}%
}
\subfloat[Foursquare]{\includegraphics[width=0.25\textwidth]{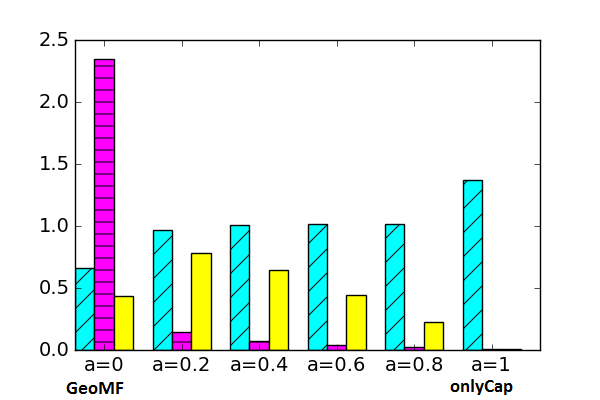}%
}
\subfloat[Gowalla]{\includegraphics[width=0.25\textwidth]{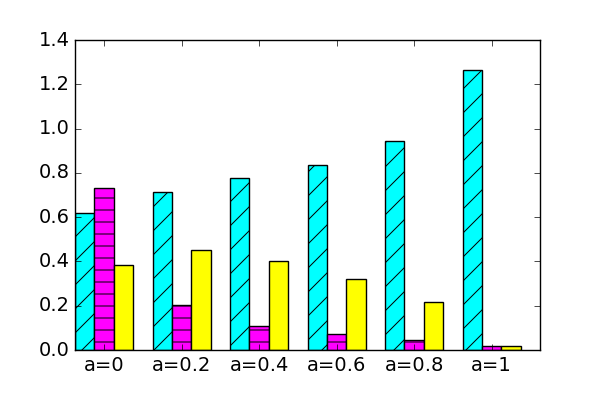}%
}
\caption{Effect of capacity trade-off parameter $\alpha$ in range $\{0, 0.2, 0.4, 0.6, 0.8, 1\}$ on \texttt{CapMF}'s performance in test RMSE, Capacity Loss and Overall Objective. As expected, the higher the $\alpha$, the higher the RMSE and the lower the Capacity Loss. 
}
\label{capparam-effect}
\end{figure*}

\subsection{Performance of \texttt{CapMF} (Rating Prediction)}
\label{subsec:q1}
The first experiment is conducted with the goal of examining the effect of the trade-off parameter $\alpha$ on \texttt{CapMF}'s performance. Recall that for \texttt{CapMF}, $\alpha$ captures the trade-off among the objective of rating prediction for whether a user will purchase (visit) an item (POI), and the objective of respecting the items' capacities. We expect that as $\alpha$ approaches 0, the algorithm will have better prediction accuracy but will be violating more the capacity constraints. In contrast, as $\alpha$ approaches 1, we expect the algorithm to  respect more the capacity constraints at the cost of worse rating prediction.

To illustrate this, we vary the trade-off parameter in the set $\{0, 0.2, 0.4, 0.6, 0.8, 1\}$. We set the surrogate loss for the capacity objective to the logistic loss, and used the `actual' capacity and propensity definitions. In Figure \ref{capparam-effect}, we report the test set metrics of RMSE, Capacity Loss and Overall objective for all four datasets. The results show that indeed the more we increase the trade-off parameter, the smaller the  capacity loss and the higher the RMSE. This is one key result of our paper, as it validates that the trade-off parameter can be used to specify to what extent the rating prediction accuracy is more/less important compared to the average capacity loss for the considered application domain. 




\subsection{Performance of \texttt{CapBPR} (Ranking)}
\label{subsec:ranking}
In the previous experiment, we had instantiated the first objective to be rating prediction accuracy, resulting in \texttt{CapMF}. Instead, here we consider the bayesian personalized ranking objective, resulting in \texttt{Cap-BPR}. Similarly with the previous experiment, we vary the trade-off parameter $\alpha$ to observe the interplay between the two objectives -- now they are 0/1 Pairwise Loss, measuring the average number of incorrectly ordered pairs, and Capacity Loss.

\begin{figure*}[!t]
\subfloat[Movielens 100K]{\includegraphics[width=0.25\textwidth]{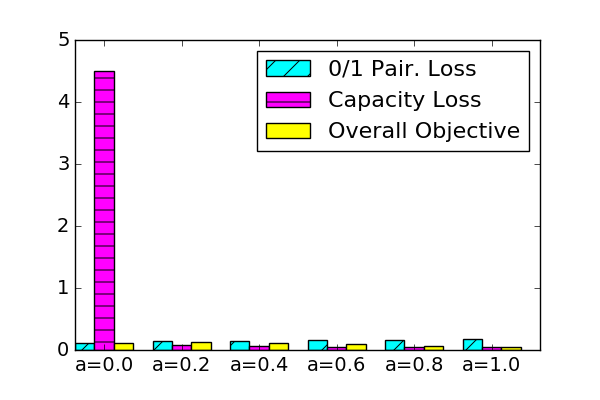}%
}
\subfloat[Foursquare]{\includegraphics[width=0.25\textwidth]{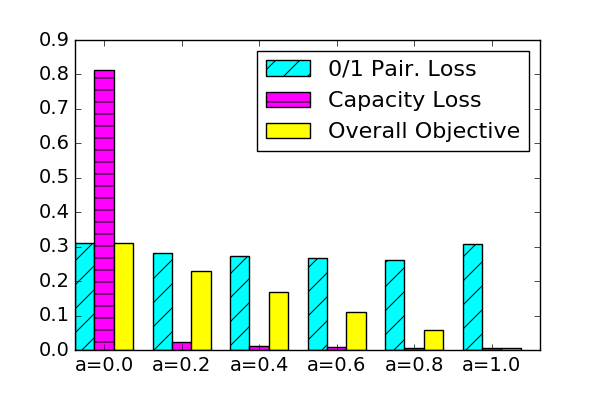}%
}
\subfloat[Movielens 100K]{\includegraphics[width=0.25\textwidth]{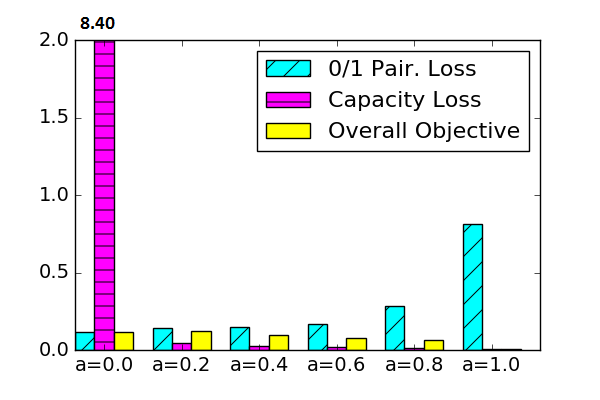}%
}
\subfloat[Foursquare]{\includegraphics[width=0.25\textwidth]{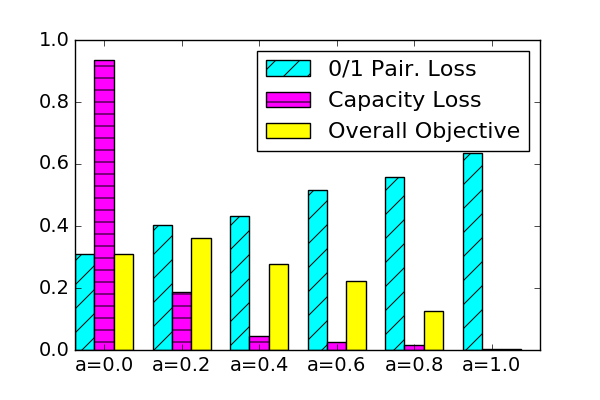}%
}
\caption{(a), (b): Actual Capacity. (c), (d): Reverse Binning Capacity. Effect of capacity trade-off parameter $\alpha$ in range $\{0, 0.2, 0.4, 0.6, 0.8, 1\}$ on \texttt{CapBPR}'s performance in test 0/1 Pairwise Loss, Capacity Loss and Overall Objective. 
}
\label{capparam-effect-rank}
\end{figure*}

\begin{figure}[ht]
\centering
\subfloat[Movielens 100K]{\includegraphics[width=0.35\textwidth]{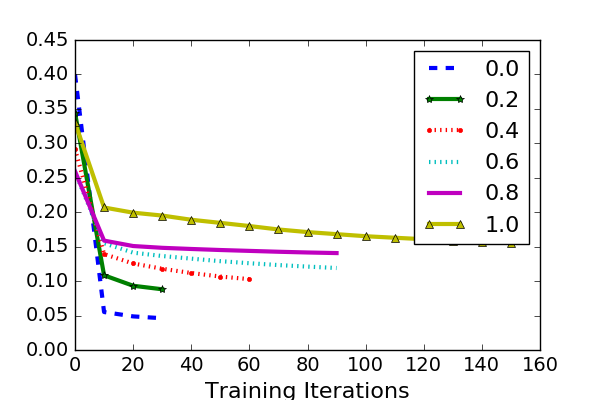}%
}
\subfloat[Foursquare]{\includegraphics[width=0.35\textwidth]{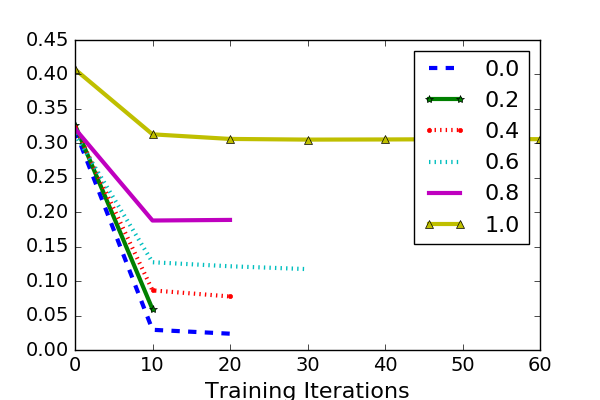}%
}
\caption{Training pairwise 0/1 loss for actual capacities.}
\label{training-progress-log}
\end{figure}

In Figures \ref{capparam-effect-rank}(a), (b), we show the results of \texttt{CapBPR}/\texttt{Cap-GeoBPR} for Movielens 100K and Foursquare, as we vary the trade-off parameter in the range of [0, 1]. Comparing these with Figures \ref{capparam-effect}(a), (c) respectively in terms of capacity loss, we see that  the ranking objective tends to result in smaller values. Also, interestingly, Figures \ref{capparam-effect-rank}(a), (b) show that while as expected the capacity loss decreases with the increase of $\alpha$, the 0/1 pairwise loss does not change much. Similar trends hold for the rest of the datasets (not shown). We found that this happens because when the capacities are analogous to usage (here we used the `actual' capacity definition), even if $\alpha$ is set to 1, i.e., not optimizing at all for ranking accuracy, the 0/1 pairwise training loss still decreases (Figure \ref{training-progress-log}). This shows that the second objective of capacity loss can help reconstruct ranking accuracy, just based on item capacities and user propensities, without any access to user-item ratings.  

To see the expected trade-off between capacity and ranking loss, we report in Figures \ref{capparam-effect-rank}(c), (d) \texttt{Cap-BPR}'s results for item capacities set inversely proportional to usage, using `reverse binning' capacities. Indeed, in this case the number of incorrectly ordered pairs increases with the increase in $\alpha$. 

\subsection{Comparison with Unconstrained Methods}
\label{subsec:q2}



Based on Figure \ref{capparam-effect}, considering the two ends of the spectrum for the value of the trade-off parameter, i.e., $\alpha=0$ and $\alpha=1$, we can compare our algorithm \texttt{CapMF} (setting $\alpha=0.2$) with the baselines \texttt{PMF}/\texttt{GeoMF} and \texttt{onlyCap} correspondingly. We observe that: (i) For item recommendation (Figure \ref{capparam-effect}(a)), \texttt{Cap-PMF} with $\alpha=0.2$ significantly improves over \texttt{PMF}'s Capacity Loss performance of 11.29 to 1.65. This happens though at the cost of deteriorated performance in terms of RMSE from .38 to .71. Similar is the trend for Movielens 1M. 
(ii) For POI recommendation (Figure \ref{capparam-effect}(c)), we observe that for Foursquare, \texttt{Cap-GeoMF}  improves over \texttt{GeoMF}'s Capacity Loss performance of 2.35 to .15, at the cost of RMSE which increases from  .66 to .97. Similar trends hold for Gowalla. 

From Figures \ref{capparam-effect-rank}(a), (b), we can compare our method \texttt{Cap-BPR} (setting $\alpha=0.2$) with the baselines of \texttt{BPR} and \texttt{GeoBPR} respectively.  We observe that for Movielens 100K, \texttt{Cap-BPR} achieves Capacity Loss of 0.08 compared to 4.51 of \texttt{BPR}, while it results in 0/1 Pairwise Loss of 0.14 compared to 0.12 (higher values are worse). Additionally, for Foursquare, \texttt{Cap-GeoBPR} achieves 0.02  Capacity Loss compared to 0.81 of \texttt{GeoBPR}, and also achieves a better 0/1 Pairwise Loss of 0.28 compared to 0.31 (the reason why was explained in Section \ref{subsec:ranking}). Similar trends hold for the rest of the datasets. 

Focusing now on the other end of the spectrum of the trade-off parameter, i.e., $\alpha = 1$, we compare \texttt{CapMF} with \texttt{onlyCap}. For example, we can see from Figure \ref{capparam-effect}(a) that for Movielens 100K, \texttt{onlyCap} (\texttt{CapMF} with $\alpha=1$) improves Capacity Loss from 1.65 to .04, but results in RMSE from .71 to 1.43. Also, \texttt{onlyCap} results in a worse 0/1 pairwise loss of 0.17, compared to 0.14 of \texttt{Cap-BPR} (Figure \ref{capparam-effect-rank}(a)). 
Similar are the trends for the rest of the datasets.

  


\subsection{Implicit versus Explicit Feedback}

\begin{figure*}[ht]
\centering
\vspace{-.5cm}
\subfloat[Movielens 100K]{\includegraphics[width=0.35\textwidth]{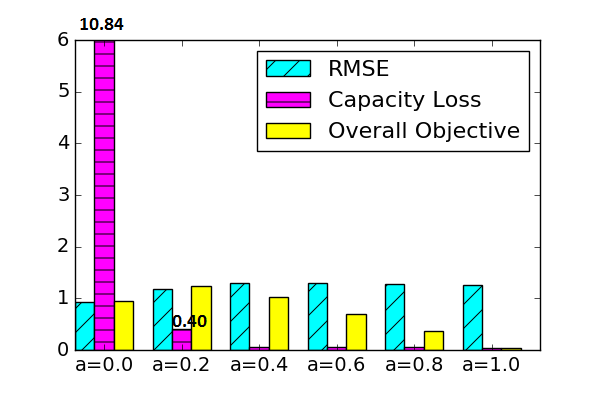}%
}
\subfloat[Movielens 1M]{\includegraphics[width=0.35\textwidth]{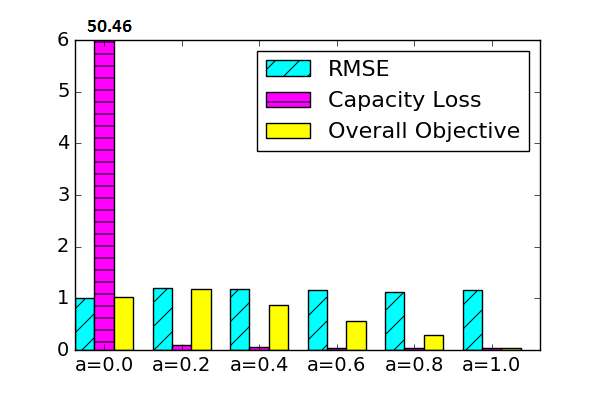}%
}
\caption{\texttt{CapMF} on explicit feedback data. }
\label{capparam-effect-explicit-square}
\end{figure*}

So far, we considered the implicit feedback version of Movielens 100K and Movielens 1M. To show that our method is also applicable when explicit feedback is available, we report in Figure \ref{capparam-effect-explicit-square} the results of \texttt{Cap-PMF} in terms of the competing objectives for the original explicit Movielens datasets. Comparing Figure \ref{capparam-effect-explicit-square}(a) with \ref{capparam-effect}(a) for Movielens 100K, and Figure \ref{capparam-effect-explicit-square}(b) with \ref{capparam-effect}(b) for Movielens 1M, we can see that although the particular values of the objectives are different, the trends are similar. Similar trends were found for \texttt{Cap-BPR}. Thus, in the sequel, we will use the implicit feedback version of the datasets unless otherwise specified. 
%

\subsection{Effect of Surrogate Loss for Capacity Term}
\label{subsec:q3}

\begin{figure*}[ht]
    \begin{minipage}{.3\textwidth}
        \centering
        \includegraphics[scale=0.35]{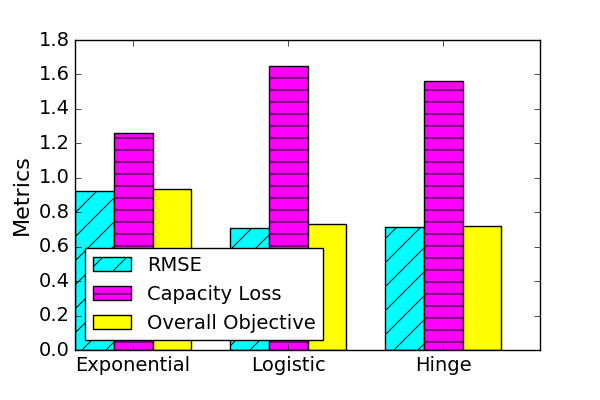}
        \caption{Movielens 100K. Effect of surrogate loss for capacity term. }
        \label{loss-effect}
    \end{minipage}\qquad%
        \begin{minipage}{0.3\textwidth}
        \centering
        \includegraphics[scale=0.35]{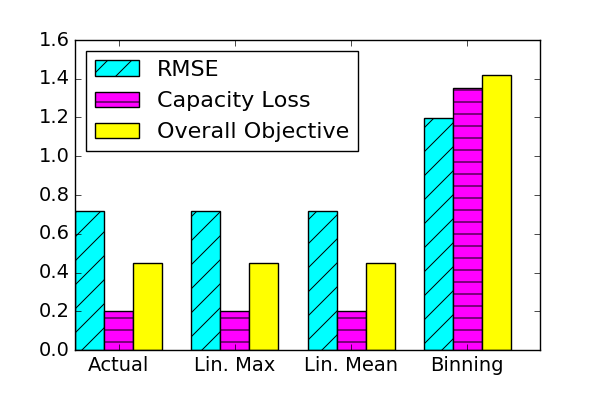}
        \caption{Gowalla. Effect of item capacities on \texttt{CapMF}. }
        \label{capacity-effect}
    \end{minipage}
    \begin{minipage}{0.3\textwidth}
        \centering
        \includegraphics[scale=0.35]{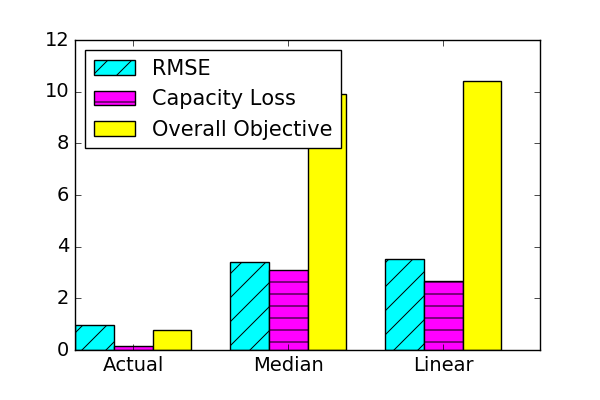}
        \caption{Foursquare. Effect of user propensities on \texttt{CapMF}.}
        \label{propensity-effect}
    \end{minipage}\qquad%
\end{figure*}

Next, we examine the effect of the choice of the surrogate loss used for the capacity objective on the algorithm's performance. Although in our experiments we have used the logistic loss, we can also use exponential or hinge loss as the surrogate loss. For this experiment, we consider rating prediction as the first objective, we set the capacity trade-off parameter to 0.2, and use the `actual' propensity and capacity definitions. Based on Figure \ref{loss-effect},  
we observe that for Movielens 100K, logistic and hinge loss result in similar performance, while exponential loss results in the highest RMSE and the smallest capacity loss. Similar are the trends for Movielens 1M (omitted). 
For Foursquare and Gowalla (omitted), hinge loss obtains the smallest capacity loss and RMSE, making it the best option for the POI datasets. 
This is the case because the exponential loss assigns very high penalties for large negative differences. Also, exponential and logistic losses assign small penalties for small negative and also positive values, while hinge does not penalize when expected usage is within the capacity constraints.   

\subsection{Effect of input Item Capacities}
\label{subsec:capacities}
Here we study the effect of different item capacities, namely `actual', `linear mean', `linear max' and `binning', as input to our algorithm. For this experiment, we set $\alpha$ to 0.2, the surrogate loss to the logistic loss, the propensities to `actual' and the first objective to rating prediction. Figure \ref{capacity-effect} compares \texttt{CapMF}'s performance for the four choices of capacity for Gowalla -- similar trends hold for the other datasets. We can see that the choices of `actual', `linear max' and `linear mean' all result in almost identical performance, whereas the choice of `binning' results in higher RMSE and Capacity Loss. We explain this result as based on Figure \ref{capacities}(d) the `binning' definition typically results in the smallest items' capacities, which means that it is more likely that the recommendations will violate the capacity constraints. Also, when setting all items' capacities uniformly to e.g. the mean actual capacity (not shown), the capacity constraints are satisfied from the first iteration.  


\subsection{Effect of input User Propensities}
\label{subsec:propensities}
In Figure \ref{propensity-effect} we study the effect of different user propensities as input to our algorithm (i.e., `actual', `median' and `linear') on \texttt{CapMF}'s performance in the Foursquare dataset. The setting is the same as the one described in  \ref{subsec:capacities}, except for the capacity definition which is set to `actual'.
We see that for `median' and `linear' choices of propensity, the values of both RMSE and Capacity Loss are higher compared to those obtained for the `actual' propensity choice. This happens because the `median' and `linear' options generally result in higher values of user propensities (Figure \ref{capacities}(a)). This leads to higher expected usage values, making it more likely to have the capacity constraints violated. 

\subsection{Effect on Top-N recommendations}
\label{subsec:q6}
\begin{figure}[h]
\vspace{-.8cm}
\centering
\subfloat[Movielens 1M]{\includegraphics[width=0.35\textwidth]{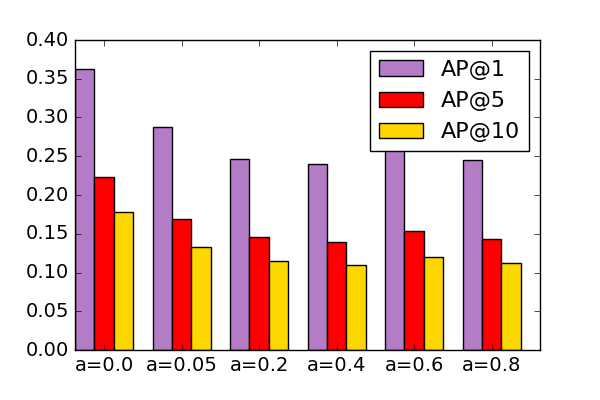}%
}
\subfloat[Foursquare]{\includegraphics[width=0.35\textwidth]{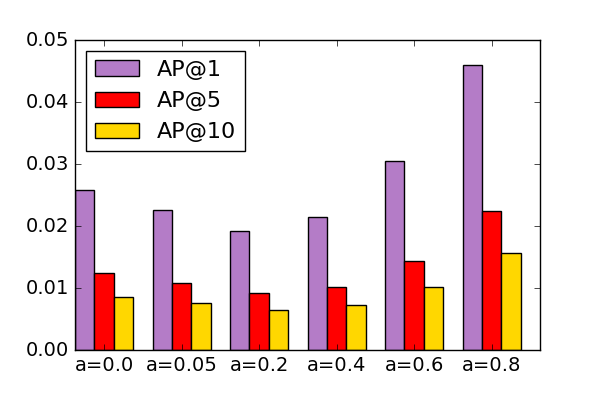}%
}
\caption{Average Precision (AP)@$\{1, 5, 10\}$.
}
\label{precisions}
\end{figure}
So far we have studied the proposed methods regarding the interplay of the two objectives (square/ranking loss versus capacity loss). Here, we evaluate the top-N recommendation quality using AP@top as our metric, varying top in $\{1, 5, 10\}$. 
Based on Figure \ref{precisions}, we see that for \texttt{CapMF}, and `actual'  capacity definition: for (a) Movielens 1M \texttt{Cap-PMF} achieves lower AP compared to \texttt{PMF}, while for (b) Foursquare, \texttt{Cap-GeoMF} achieves higher AP compered to \texttt{GeoMF}. 
While we increase the top, AP@top decreases. 
Due to space constraints, we omitted AP@top results for all datasets and capacity definitions. Overall we noticed that: (i) Our method with rating prediction loss tends to outperform \texttt{PMF}/\texttt{GeoMF} when the capacities are inversely proportional or irrespective of usage. (ii) In contrast, for ranking loss, our methods tend to outperform \texttt{BPR}/\texttt{Cap-BPR} when capacities are proportional or irrespective of usage. 

\subsection{Comparison with Post-Processing Method}
\label{subsec:baseline}

Last, we compare our proposed approach with a simple post-processing baseline, outlined in Algorithm \ref{algorithm:1}. By design, such a baseline will never violate the capacity constraints present. 

\begin{algorithm}
\caption{Post-Processing Baseline}
\begin{algorithmic}[1]
\REQUIRE $\forall j~ c_j$, $\forall (i, j)~~ \hat{r}_{ij} $ from unconstrained method (PMF, GeoMF, BPR)
\STATE $\forall \text{ item } j$ sort users $i$ based on $\hat{r}_{ij}$
\STATE Recommend item $j$ only to the top $c_j$ users \\ 
\RETURN For every user $i$, give top-N list, sorting the recommended items based on $\hat{r}_{ij}$ 
\end{algorithmic}
\label{algorithm:1}
\end{algorithm}

For the purposes of the novel setting of recommendation under capacity constraints, here we introduce two new metrics: 
\begin{align}
& \label{eq:weigthedap} \text{Weighted AP @ top (WAP@top)} = \frac{\sum_{i = 1}^{M} p_i \text{AP}_i @ top}{\sum_{i=1}^{M} p_i}. \\
& \label{eq:weigthedviol}WMCV@top = \frac{1}{N} \sum_{j=1}^{N} \mathbf{1}\left[\sum_{i\in\text{Re}^{\text{top}}(j)} p_i \geq c_j \right], 
\end{align} where $\text{AP}_i$ is user $i$'s AP, $\text{Re}^{\text{top}}(j)$ is the set of users who got $j$ in their top list, and WMCV stands for Weighted Mean Capacity Violation.  WAP@top measures top-N recommendation quality  taking into account the users' propensities. WMCV@top captures by how much  on average the items' capacities are violated, measuring the extra number of users recommended the item in their top-N list weighted by their propensities. 
\begin{figure}[!h]
\vspace{-0.5cm}
\centering
\subfloat[Cap-BPR, actual]{\includegraphics[width=0.45\textwidth]{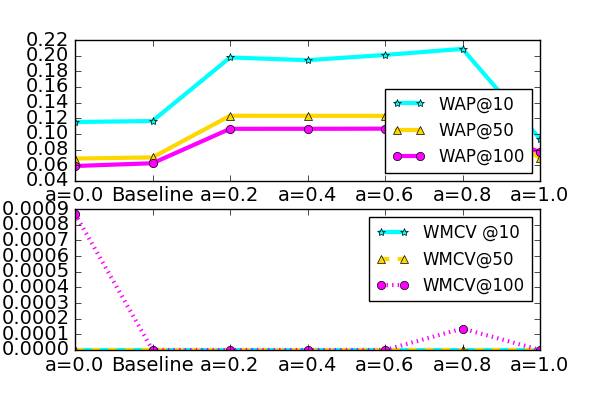}%
}
\subfloat[Cap-BPR, reverse binning]{\includegraphics[width=0.45\textwidth]{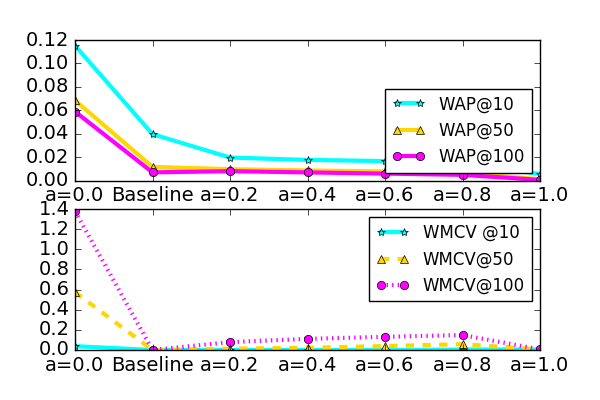}%
}
\caption{Movielens 100K, explicit: Comparison with baselines in WAP, WMCV. 
}
\label{baselin-exp-rank}
\end{figure}

Figure \ref{baselin-exp-rank} compares \texttt{Cap-BPR} with the proposed post-processing baseline for (a) `actual' and (b) `reverse binning' capacities for the explicit version of Movielens 100K. We can see that for `actual' capacities (as well as `binning', `uniform') , \texttt{Cap-BPR} not only outperforms in terms of WAP@top the baseline of Algorithm 1 but also surpasses \texttt{BPR}. In contrast, for `reverse binning', one should employ the simple baseline to achieve better WAP@top. When the first objective is rating prediction (omitted), the trend is different: \texttt{CapMF} outperforms the baseline when capacities are irrespective or reversely proportional to usage, whereas when the capacities are proportional to usage, the baseline is better. Similar trends hold for the non-weighted version of Average Precision. For POI datasets (omitted), \texttt{Cap-GeoMF} outperforms both the baseline and \texttt{GeoMF} either when the capacities are proportional or inversely proportional to usage. 
In terms of WMCV@ top, for larger values of top, our method outperforms the unconstrained methods. For smaller values of top, the unconstrained methods already achieve WMCV close to zero. Interestingly, an increase in $\alpha$ does not always translate to smaller WMCV@top.

\section{Conclusions}
\label{sec:concl}
We have presented a novel approach for providing recommendations that satisfy capacity constraints. 
We have demonstrated how this generic approach can be applied to three state-of-the-art models, PMF \cite{salakhutdinov2011probabilistic}, GeoMF \cite{lian2014geomf}, and BPR \cite{rendle2009bpr}, so that the items' expected usage respects the corresponding capacities. We have empirically shown the trade-off among rating prediction / ranking accuracy and capacity loss, and have illustrated the effect of various parameters.  
We have shown that our method achieves comparable top-N recommendation performance with the state-of-the-art while achieving lower capacity loss,  making it applicable in real-world recommendation scenarios where capacity constraints are present.

\vspace*{3mm}
{\bf Acknowledgements:} The research was supported by NSF grants IIS-1563950, IIS-1447566, IIS-1422557, CCF-1451986, CNS-
1314560, IIS-0953274, IIS-1029711, and by NASA grant NNX12AQ39A. Arindam Banerjee acknowledges support from IBM, Yahoo and Adobe Research. Access to computing facilities were provided by the University of Minnesota Supercomputing Institute (MSI).


\begin{thebibliography}{10}

\bibitem{agarwal2009explore}
Deepak Agarwal, Bee-Chung Chen, and Pradheep Elango.
\newblock Explore/exploit schemes for web content optimization.
\newblock In {\em Proceedings of the 2009 9th IEEE International Conference on
  Data Mining}, pages 1--10. IEEE, 2009.

\bibitem{agarwal2011click}
Deepak Agarwal, Bee-Chung Chen, Pradheep Elango, and Xuanhui Wang.
\newblock Click shaping to optimize multiple objectives.
\newblock In {\em Proceedings of the 17th ACM SIGKDD international conference
  on Knowledge discovery and data mining}, pages 132--140. ACM, 2011.

\bibitem{agarwal2012personalized}
Deepak Agarwal, Bee-Chung Chen, Pradheep Elango, and Xuanhui Wang.
\newblock Personalized click shaping through lagrangian duality for online
  recommendation.
\newblock In {\em Proceedings of the 35th international ACM SIGIR conference on
  Research and development in information retrieval}, pages 485--494. ACM,
  2012.

\bibitem{christakopoulou2015collaborative}
Konstantina Christakopoulou and Arindam Banerjee.
\newblock Collaborative ranking with a push at the top.
\newblock In {\em Proceedings of the 24th International Conference on World
  Wide Web}, pages 205--215. ACM, 2015.

\bibitem{duchi2011adaptive}
John Duchi, Elad Hazan, and Yoram Singer.
\newblock Adaptive subgradient methods for online learning and stochastic
  optimization.
\newblock {\em Journal of Machine Learning Research}, 12(Jul):2121--2159, 2011.

\bibitem{gao2013exploring}
Huiji Gao, Jiliang Tang, Xia Hu, and Huan Liu.
\newblock Exploring temporal effects for location recommendation on
  location-based social networks.
\newblock In {\em Proceedings of the 7th ACM conference on Recommender
  systems}, pages 93--100. ACM, 2013.

\bibitem{Herbrich16}
Ralf Herbrich.
\newblock Learning sparse models at scale.
\newblock In {\em Proceedings of the 22nd {ACM} {SIGKDD} International
  Conference on Knowledge Discovery and Data Mining, San Francisco, CA, USA,
  August 13-17, 2016}, page 407, 2016.

\bibitem{hu2008collaborative}
Yifan Hu, Yehuda Koren, and Chris Volinsky.
\newblock Collaborative filtering for implicit feedback datasets.
\newblock In {\em Proceedings of the 2008 8th IEEE International Conference on
  Data Mining}, pages 263--272. IEEE, 2008.

\bibitem{jambor2010optimizing}
Tamas Jambor and Jun Wang.
\newblock Optimizing multiple objectives in collaborative filtering.
\newblock In {\em Proceedings of the 4th ACM conference on Recommender
  systems}, pages 55--62. ACM, 2010.

\bibitem{johnson2014logistic}
Christopher~C Johnson.
\newblock Logistic matrix factorization for implicit feedback data.
\newblock In {\em NIPS 2014 Workshop on Distributed Machine Learning and Matrix
  Computations}, 2014.

\bibitem{karatzas2012brownian}
Ioannis Karatzas and Steven Shreve.
\newblock {\em Brownian motion and stochastic calculus}, volume 113.
\newblock Springer Science \& Business Media, 2012.

\bibitem{koren2009matrix}
Yehuda Koren, Robert Bell, Chris Volinsky, et~al.
\newblock Matrix factorization techniques for recommender systems.
\newblock {\em Computer}, 42(8):30--37, 2009.

\bibitem{lipoint}
Huayu Li, Yong Ge, and Hengshu Zhu.
\newblock Point-of-interest recommendations: Learning potential check-ins from
  friends.
\newblock In {\em Proceedings of the 22nd ACM SIGKDD international conference
  on Knowledge discovery and data mining}, pages 975--984. ACM, 2016.

\bibitem{li2015rank}
Xutao Li, Gao Cong, Xiao-Li Li, Tuan-Anh~Nguyen Pham, and Shonali Krishnaswamy.
\newblock Rank-geofm: a ranking based geographical factorization method for
  point of interest recommendation.
\newblock In {\em Proceedings of the 38th International ACM SIGIR Conference on
  Research and Development in Information Retrieval}, pages 433--442. ACM,
  2015.

\bibitem{lian2014geomf}
Defu Lian, Cong Zhao, Xing Xie, Guangzhong Sun, Enhong Chen, and Yong Rui.
\newblock Geomf: joint geographical modeling and matrix factorization for
  point-of-interest recommendation.
\newblock In {\em Proceedings of the 20th ACM SIGKDD international conference
  on Knowledge discovery and data mining}, pages 831--840. ACM, 2014.

\bibitem{liang2016modeling}
Dawen Liang, Laurent Charlin, James McInerney, and David~M Blei.
\newblock Modeling user exposure in recommendation.
\newblock In {\em Proceedings of the 25th International Conference on World
  Wide Web}, pages 951--961. International World Wide Web Conferences Steering
  Committee, 2016.

\bibitem{mairal2010online}
Julien Mairal, Francis Bach, Jean Ponce, and Guillermo Sapiro.
\newblock Online learning for matrix factorization and sparse coding.
\newblock {\em Journal of Machine Learning Research}, 11(Jan):19--60, 2010.

\bibitem{pan2008one}
Rong Pan, Yunhong Zhou, Bin Cao, Nathan~N Liu, Rajan Lukose, Martin Scholz, and
  Qiang Yang.
\newblock One-class collaborative filtering.
\newblock In {\em Data Mining, 2008. ICDM'08. Eighth IEEE International
  Conference on}, pages 502--511. IEEE, 2008.

\bibitem{rendle2009bpr}
Steffen Rendle, Christoph Freudenthaler, Zeno Gantner, and Lars Schmidt-Thieme.
\newblock Bpr: Bayesian personalized ranking from implicit feedback.
\newblock In {\em Proceedings of the 25th conference on uncertainty in
  artificial intelligence}, pages 452--461. AUAI Press, 2009.

\bibitem{salakhutdinov2011probabilistic}
Ruslan Salakhutdinov and Andriy Mnih.
\newblock Probabilistic matrix factorization.
\newblock In {\em Proceedings of the 20th Neural Information Processing
  Systems}, volume~20, pages 1257--1264, 2007.

\bibitem{schnabel2016recommendations}
Tobias Schnabel, Adith Swaminathan, Ashudeep Singh, Navin Chandak, and Thorsten
  Joachims.
\newblock Recommendations as treatments: Debiasing learning and evaluation.
\newblock {\em arXiv preprint arXiv:1602.05352}, 2016.

\bibitem{shan2010generalized}
Hanhuai Shan and Arindam Banerjee.
\newblock Generalized probabilistic matrix factorizations for collaborative
  filtering.
\newblock In {\em Proceedings of the 2010 IEEE International Conference on Data
  Mining}, pages 1025--1030. IEEE, 2010.

\bibitem{shi2012climf}
Yue Shi, Alexandros Karatzoglou, Linas Baltrunas, Martha Larson, Nuria Oliver,
  and Alan Hanjalic.
\newblock Climf: learning to maximize reciprocal rank with collaborative
  less-is-more filtering.
\newblock In {\em Proceedings of the 6th ACM conference on Recommender
  systems}, pages 139--146. ACM, 2012.

\bibitem{svore2011learning}
Krysta~M Svore, Maksims~N Volkovs, and Christopher~JC Burges.
\newblock Learning to rank with multiple objective functions.
\newblock In {\em Proceedings of the 20th international conference on World
  wide web}, pages 367--376. ACM, 2011.

\bibitem{yu2015survey}
Yonghong Yu and Xingguo Chen.
\newblock A survey of point-of-interest recommendation in location-based social
  networks.
\newblock In {\em Workshops at the 29th AAAI Conference on Artificial
  Intelligence}, 2015.

\bibitem{zhang2013igslr}
Jia-Dong Zhang and Chi-Yin Chow.
\newblock igslr: personalized geo-social location recommendation: a kernel
  density estimation approach.
\newblock In {\em Proceedings of the 21st ACM SIGSPATIAL International
  Conference on Advances in Geographic Information Systems}, pages 334--343.
  ACM, 2013.

\end{thebibliography}
\bibliographystyle{abbrv}

\end{document}